\pdfoutput=1
\documentclass[11pt]{article}

\usepackage{ACL2023}

\usepackage{times}
\usepackage{latexsym}

\usepackage[T1]{fontenc}
\usepackage[utf8]{inputenc}

\usepackage{microtype}

\usepackage{inconsolata}

\title{Span-Selective Linear Attention Transformers for Effective and Robust Schema-Guided Dialogue State Tracking}

\author{Bj\"orn Bebensee \quad Haejun Lee \\
  Samsung Research \\
  \texttt{\{b.bebensee,haejun82.lee\}@samsung.com} \\}%\And
\usepackage{amsmath}
\usepackage{amssymb}
\usepackage{booktabs}
\usepackage{placeins}
\usepackage{mathtools}
\usepackage{adjustbox}
\usepackage{pifont}
\usepackage{todonotes}
\usepackage{bbm}

\newcommand{\cmark}{\ding{51}}%
\newcommand{\xmark}{\ding{55}}%

\begin{document}

\maketitle

\begin{abstract}
In schema-guided dialogue state tracking models estimate the current state of a conversation using natural language descriptions of the service schema for generalization to unseen services. Prior generative approaches which decode slot values sequentially do not generalize well to variations in schema, while discriminative approaches separately encode history and schema and fail to account for inter-slot and intent-slot dependencies. We introduce SPLAT, a novel architecture which achieves better generalization and efficiency than prior approaches by constraining outputs to a limited prediction space. At the same time, our model allows for rich attention among descriptions and history while keeping computation costs constrained by incorporating linear-time attention. 
We demonstrate the effectiveness of our model on the Schema-Guided Dialogue (SGD) and MultiWOZ datasets. Our approach significantly improves upon existing models achieving 85.3 JGA on the SGD dataset. Further, we show increased robustness on the SGD-X benchmark: our model outperforms the more than 30$\times$ larger D3ST-XXL model by 5.0 points.
\end{abstract}

\FloatBarrier

\section{Introduction}
Dialogue State Tracking (DST) refers to the task of estimating and tracking the dialogue state consisting of the user's current intent and set of slot-value pairs throughout the dialogue \citep{williams-etal-2013-dialog}. Traditional approaches to DST assume a fixed ontology and learn a classifier for each slot \citep{chao2019bertdst}. However, in real-world applications services can be added or removed requiring the model to be re-trained each time the ontology changes. Recently more flexible schema-guided approaches which take as input natural language descriptions of all available intents and slots and thus can be applied zero-shot to new services have been gaining popularity \citep{rastogi2020towards,feng2021sequence,zhao2022description,gupta2022show}.

\begin{figure}
    \centering
    \includegraphics[width=\linewidth]{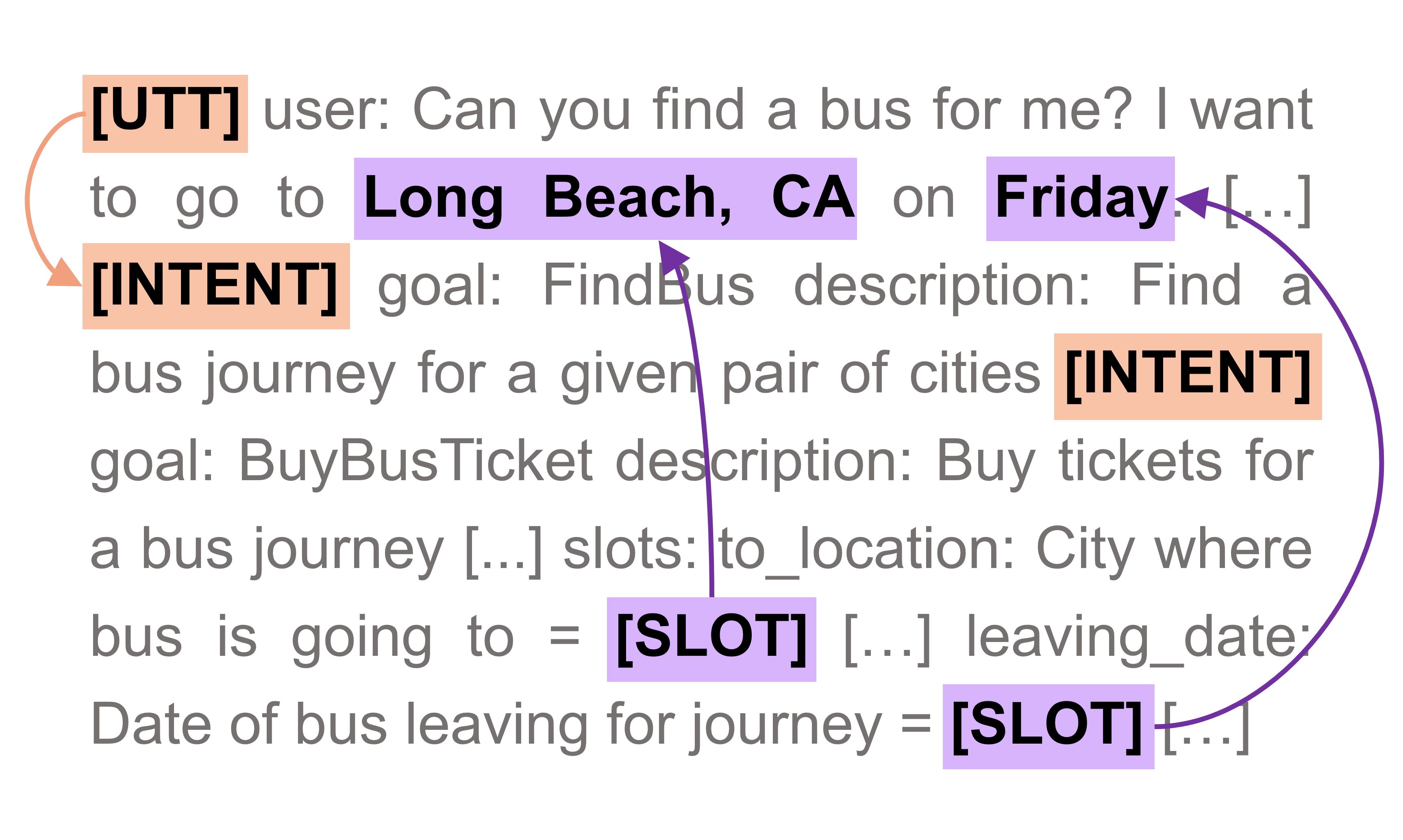}
    \caption{Span selection for schema-guided dialogue in practice. \texttt{[SLOT]} encodes the semantics of the natural language description of ``\texttt{to\_location}'' and is matched with the span representation of ``Long Beach, CA''. Similarly \texttt{[UTT]} encodes the semantics of the current utterance and is matched with the target \texttt{[INTENT]} encoding.}
    \label{fig:pointers}
\end{figure}

Discriminative DST models are based on machine reading comprehension (MRC) methods, meaning they extract and fill in non-categorical slot values directly from the user utterances \citep{chao2019bertdst,ruan2020fine,zhang2021sgd}. We use the terms discriminative and extractive interchangeably when referring to these methods. Generative DST models leverage seq2seq language models which conditioned on the dialog history and a prompt learn to sequentially generate the appropriate slot values. Prior generative methods do not generalize well to variations in schema \citep{lee2021dialogue, sgdx, zhao2022description} whereas discriminative methods separately encode history and schema and fail to account for inter-slot and intent-slot dependencies.

\begin{figure*}
    \centering
    \includegraphics[width=\textwidth]{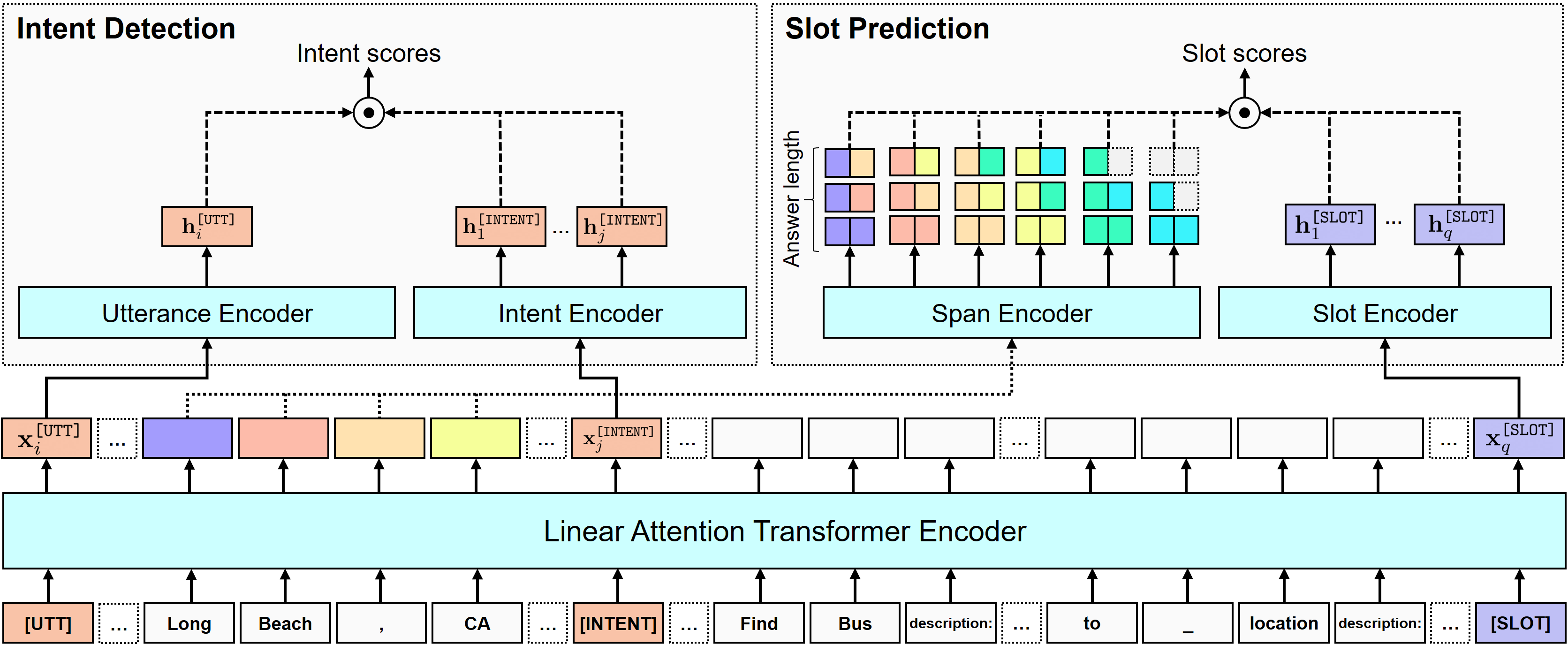}
    \caption{An overview over the SPLAT model architecture. Intent scores are computed using the utterance representation $\mathbf{h}_i^\texttt{[UTT]}$ and intent representations $\mathbf{h}_k^\texttt{[INTENT]}$. A span encoder computes span representations $\mathbf{h}_{mn}^\texttt{SPAN}$ for all spans $x_m, \ldots, x_n$. The target span is selected by matching the slot query $\mathbf{h}_q^\texttt{[SLOT]}$ to the target span $\mathbf{h}_{ij}^\texttt{SPAN}$.}
    \label{fig:architecture}
\end{figure*}

In this work we introduce the \textbf{SP}an-Selective \textbf{L}inear \textbf{A}ttention \textbf{T}ransformer, short SPLAT, a novel architecture designed to achieve better generalization, robustness and efficiency in DST than existing approaches. SPLAT is fully extractive and, unlike prior generative approaches, constrains the output space to only those values contained in the input sequence. Figure \ref{fig:pointers} shows an example of the key idea behind our approach. We jointly encode the natural language schema and full dialogue history allowing for a more expressive contextualization. Spans in the input are represented by aggregating semantics of each individual span into a single representation vector. Then we take a contrastive query-based pointer network approach \citep{vinyals2015pointer} to match special query tokens to the target slot value's learned span representation in a single pass. 

Our main contributions are as follows:
\begin{itemize}
    \item We propose novel span-selective prediction layers for DST which provide better generalization and efficiency by limiting the prediction space and inferring all predictions in parallel. We achieve state-of-the-art performance on the SGD-X benchmark outperforming the 30$\times$ larger D3ST by 5.0 points.
    \item We adopt a Linear Attention Transformer which allows more expressive contextualization of the dialogue schema and dialogue history with constrained prediction time. We show our model already outperforms other models with similar parameter budgets even without other modules we propose in Table~\ref{table:sgd-results} and \ref{table:ablation-results}.
    \item We pre-train SPLAT for better span representations with a recurrent span selection objective yielding significant further span prediction performance gains of up to 1.5 points.
\end{itemize}
 
\section{Approach}

\subsection{Task Formulation}

For a given dialog of $T$ turns let $U$ describe the set of utterances in the dialog history $U = \{ u_1, \ldots, u_T \}$. Each $u_i$ can represent either a user or a system utterance. The system is providing some service to the user defined by a service schema $S$. The service schema consists of a set of intents $I = \{ i_1, \ldots, i_K \}$ and their intent descriptions $D^{\text{intent}} = \{ d^{\mathrlap{{\text{intent}}}}_1 \ , \ldots, d^{\text{intent}}_K \}$ as well as a set of slots $S = \{ s_1, \ldots, s_L\}$ and their slot descriptions $D^{\text{slot}} = \{ d^{\mathrlap{\text{slot}}}_1 \ , \ldots, d^{\text{slot}}_L \}$.

In practice we prepend each $u_i$ with the speaker name ($\texttt{user}$ or $\texttt{system}$) and a special utterance query token $\texttt{[UTT]}$ which will serve as the encoding of the system-user utterance pair.

Each $d^{\text{slot}}_i$ consists of the slot name, a natural language description of the semantics of the slot and for categorical values an enumeration of all possible values this slot can assume. We also append a special slot query embedding token $\texttt{[SLOT]}$ which serves as the slot encoding.

Some slot values are shared across all slots and their representation can be modeled jointly. Unless denoted otherwise these shared target values $T$ are special tokens $\texttt{[NONE]}$ and $\texttt{[DONTCARE]}$ which correspond to the "none" and "dontcare" slot values in SGD and MultiWOZ.

\subsection{Joint Encoding with Linear Attention}

\paragraph{Linear Attention Transformers.} In order to better capture the semantics of the input and to allow for a longer context as well as all the relevant schema descriptions to be encoded jointly we use a Transformer \citep{vaswani2017attention} with linear-time attention. Instead of computing the full attention matrix as the original Transformer does, its linear attention variants compute either an approximation of it \citep{choromanski2021rethinking} or only compute full attention for a fixed context window of size $w$ around the current token and additional $n_\text{global}$ global tokens, thus lowering the complexity of the attention computation from $\mathcal{O}(n^2)$ for a sequence of length $n$ to $\mathcal{O}(w + n_\text{global})$ \citep{beltagy2020longformer, zaheer2020big}.

We focus on the windowed variant and incorporate it to DST. We denote the Linear Attention Transformer with selective global attention parametrized by $\theta$ with input sequence $\mathcal{I}$ and its subset of global input tokens $\mathcal{G} \subseteq \mathcal{I}$, i.e. inputs corresponding to tokens at positions that are attended using the global attention mechanism, as $\texttt{LAT}(\mathcal{I};\mathcal{G};\theta)$. While we choose the Longformer \citep{beltagy2020longformer} for our implementation, in practice any variants with windowed and global attention can be used instead.

\paragraph{Joint encoding.} The full input sequence of length $N$ is given as the concatenation of its components. We define the set of globally-attended tokens as the union of sets of tokens corresponding to the intent descriptions $D^{\text{intent}}$, the slot descriptions $D^{\text{slot}}$, and the shared target values $T$. Then, the joint encoding of $N$ hidden states is obtained as the output of the last Transformer layer as
\begin{align}
    \mathcal{I} &= \texttt{[CLS]} \ U \ \texttt{[SEP]} \ T \ D^{\text{intent}} \ D^{\text{slot}} \ \texttt{[SEP]} \nonumber \\
     \mathcal{G} &= T \cup D^{\text{intent}} \cup D^{\text{slot}} \nonumber \\
     E &= \texttt{LAT}(\mathcal{I}; \mathcal{G}; \theta).
\end{align}

\subsection{Intent Classification}

Let $\mathbf{x}_i^\texttt{[UTT]}$ denote the representation of the encoded $\texttt{[UTT]}$ token corresponding to the $i$-th turn. Given the encoded sequence $E$, we obtain the final utterance representations by feeding $\mathbf{x}_i^\texttt{[UTT]}$ into the utterance encoder. Similarly for each intent $I = \{i_1, \ldots, i_t \}$ and its respective $\texttt{[INTENT]}$ token, we obtain final intent representations using the intent encoder:
\begin{align}
    \mathbf{h}_i^\texttt{[UTT]} &= \mathrm{LN}(\mathrm{FFN}(\mathbf{x}_i^\texttt{[UTT]})) \nonumber \\
    \mathbf{h}_j^\texttt{[INTENT]} &= \mathrm{LN}(\mathrm{FFN}(\mathbf{x}_j^\texttt{[INTENT]}))
\end{align}

\noindent Here LN refers to a LayerNorm and FFN to a feed-forward network. 

We maximize the dot product similarity between each utterance representation and the ground truth active intent's representation via cross-entropy:
\begin{align}
    \mathrm{score}_{i \rightarrow j} &= \mathrm{sim}(\mathbf{h}_i^\texttt{[UTT]}, \mathbf{h}_j^\texttt{[INTENT]}) \nonumber \\
    \mathcal{L}_\text{intent} = -\frac{1}{T} &\sum_{i=1}^T \log \frac{\exp(\mathrm{score}_{i \rightarrow j})}{\sum_{k=1}^K \exp(\mathrm{score}_{i \rightarrow k})} \cdot \mathbbm{1}_\text{GT}
\end{align}

\noindent where $K$ is the number of intents and $\mathbbm{1}_\text{GT}$ is an indicator function which equals 1 if and only if $j$ is the ground truth matching $i$.

\subsection{Span Pointer Module}

We introduce a novel Span Pointer Module which computes span representations via a span encoder and extracts slot values by matching slot queries via a similarity-based span pointing mechanism~\citep{vinyals2015pointer}.

First, for any given span of token representations $\mathbf{x}_i, \ldots, \mathbf{x}_j$ in the joint encoding $E$ we obtain the span representation $\mathbf{h}_{ij}^\texttt{SPAN}$ by concatenating the span's first and last token representation and feeding them into a 2-layer feed-forward span encoder \citep{joshi2020spanbert}:
\begin{gather}
    \mathbf{y}_{ij} = [\mathbf{x}_i ; \mathbf{x}_j]  \nonumber \\
    \mathbf{h}_{ij}^\texttt{SPAN} = \mathrm{LN}(\mathrm{FFN}_\mathrm{GeLU}(\mathbf{y}_{ij})) \times \texttt{n\_layers}
\end{gather}

Similarly, for each slot token representation $\mathbf{x}^\texttt{[SLOT]}$ in $E$ we compute a slot query representation $\mathbf{h}^\texttt{[SLOT]}$ with a 2-layer feed-forward slot encoder:
\begin{equation}
    \mathbf{h}^\texttt{[SLOT]} = \mathrm{LN}(\mathrm{FFN}_\mathrm{GeLU}(\mathbf{x}^{\texttt{[SLOT]}})) \times \texttt{n\_layers}
\end{equation}

Given slots $S = \{s_1, \ldots, s_L \}$ and corresponding slot query representations $\mathbf{h}_1^{\mathrlap{\texttt{[SLOT]}}} \ , \ldots, \mathbf{h}_L^\texttt{[SLOT]}$ we score candidate target spans by dot product similarity of the slot queries with their span representations. That is, for each slot query $q$ with ground truth target span $x_i, \ldots, x_j$ we maximize $\mathrm{sim}(\mathbf{h}_q^\texttt{[SLOT]}, \mathbf{h}_{ij}^{\texttt{SPAN}})$ by cross-entropy. The loss function is given by
\begin{gather}
    \mathrm{score}_{q \rightarrow ij} = \mathrm{sim}(\mathbf{h}_q^{\texttt{[SLOT]}}, \mathbf{h}_{ij}^{\texttt{SPAN}}) \nonumber \\
    \mathcal{L}_\text{slot} = -\frac{1}{L} \sum_{q=1}^{L} \log \frac{\exp(\mathrm{score}_{q \rightarrow ij})}{\sum_{k=1}^K \exp(\mathrm{score}_{q \rightarrow k})} \cdot \mathbbm{1}_\text{GT}
    \label{eq:slot_loss}
\end{gather}
where $L$ is the number of slots and $K$ is the number of spans. $\mathrm{sim}(\mathbf{h}_q^{\texttt{[SLOT]}}, \mathbf{h}_{ij}^{\texttt{SPAN}})$ denotes the similarity between the $q$-th slot query representation and the span representation of its ground truth slot value.

It is computationally too expensive to compute span representations for all possible spans. In practice however the length of slot values rarely exceeds some $L_\text{ans}$. Thus, we limit the maximum span length to $L_\text{ans}$ and do not compute scores for spans longer than this threshold. This gives us a total number of $N \cdot L_\text{ans}$ candidate spans.

\paragraph{Joint optimization.} We optimize the intent and slot losses jointly via the following objective:

\begin{equation}
    \mathcal{L} = \frac{\mathcal{L}_\text{slot} + \mathcal{L}_\text{intent}}{2}
\end{equation}

\subsection{Pre-Training via Recurrent Span Selection}

Since the span pointer module relies on span embedding similarity for slot classification we believe it is crucial to learn good and robust span representations. In order to improve span representations for down-stream applications to DST we pre-train SPLAT in a self-supervised manner using a modified recurrent span selection objective \citep{ram2021fewshot}.

Given an input text $\mathcal{I}$ let $\mathcal{R} = \{ \mathcal{R}_1, \ldots, \mathcal{R}_a \} $ be the clusters of identical spans that occur more than once. Following \citet{ram2021fewshot} we randomly select a subset $\mathcal{M} \subseteq \mathcal{R}$ of $J$ recurring spans such that the number of their occurrences sums up to a maximum of 30 occurrences. Then, for each selected cluster of recurring spans $\mathcal{M}_j$ we randomly replace all but one occurrence with the query token $\texttt{[SLOT]}$.

The slot query tokens act as the queries while the respective unmasked span occurrences act as the targets. Unlike the original recurrent span selection objective we do not use separate start and end pointers for the target spans but instead use our Span Pointer Module to learn a single representation for each target span.

We pre-train SPLAT to maximize the dot product similarity between the query token and the unmasked target span representation. The loss for the $j$-th cluster of identical masked spans is given by Equation (\ref{eq:slot_loss}) and the total loss is given as the sum of losses of over all clusters.

Effectively each sentence containing a masked occurrence of the span acts as the span description while the target span acts as the span value. This can be seen as analogous to slot descriptions and slot values in DST.

\section{Experimental Setup}

We describe our experimental setup including datasets used for pre-training and evaluation, implementation details, baselines and evaluation metrics in detail below.

\subsection{Benchmark Datasets}

We conduct experiments on the Schema-Guided Dialogue (SGD) \citep{rastogi2020towards}, SGD-X \citep{sgdx} and MultiWOZ 2.2 \citep{multiwoz2.2} datasets.

\paragraph{Schema-Guided Dialogue.} Unlike other task-oriented dialogue datasets which assume a single, fixed ontology at training and test time the SGD dataset includes new and unseen slots and services in the test set. This allows us to not only measure DST performance but also zero-shot generalization to unseen services. The dataset includes natural language descriptions for all intents and slots in its schema. We follow the standard evaluation setting and data split suggested by the authors.
 
\paragraph{SGD-X.} The SGD-X benchmark is an extension of the SGD dataset which provides five additional schema variants of different linguistic styles which increasingly diverge in style from the original schema with $v_1$ being most similar and $v_5$ least similar. We can evaluate our model's robustness to variations in schema descriptions by training our model on SGD and comparing evaluation results using the different included schema variants.

\paragraph{MultiWOZ.} The MultiWOZ dataset is set of human-human dialogues collected in the Wizard-of-OZ setup. Unlike in SGD the ontology is fixed and there are no unseen services at test time. There are multiple updated versions of the original MultiWOZ dataset \citep{multiwoz2.0}: MultiWOZ 2.1 \citep{multiwoz2.1} and MultiWOZ 2.2 \citep{multiwoz2.2} fix annotation errors of previous versions, MultiWOZ 2.3 \citep{multiwoz2.3} is based on version 2.1 and adds co-reference annotations, MultiWOZ 2.4 \citep{multiwoz2.4} is also based on version 2.1 and includes test set corrections. However, MultiWOZ 2.2 is the only version of the dataset which includes a fully defined schema matching the ontology. We therefore choose the MultiWOZ 2.2 dataset for our experiments. We follow the standard evaluation setting and data split.

\subsection{Evaluation Metrics}
In line with prior work~\cite{rastogi2020towards} we evaluate our approach according to the following two metrics. \\
\textbf{Intent Accuracy:} For intent detection the intent accuracy describes the fraction of turns for which the active intent has been correctly inferred. \\
\textbf{Joint Goal Accuracy (JGA):} For slot prediction JGA describes the fraction of turns for which all slot values have been predicted correctly. Following the evaluation setting from each dataset we use a fuzzy matching score for slot values in SGD and exact match in MultiWOZ.

\subsection{Implementation Details}
We base our implementation on the Longformer code included in the HuggingFace Transformers library \citep{huggingface-transformers} and continue training from the base model (110M parameters) and large model (340M parameters) checkpoints. We keep the default Longformer hyperparameters in place, in particular we keep the attention window size set to 512. The maximum sequence length is 4096.
During pre-training we train the base model for a total of 850k training steps and the large model for 800k training steps. During fine-tuning we train all models for a single run of 10 epochs and choose the model with the highest joint goal accuracy on the development set. We use the Adam optimizer \citep{kingma2014adam} with a maximum learning rate of $10^{-5}$ which is warmed up for the first 10\% of steps and subsequently decays linearly. We set the batch size to 32 for base models and to 16 for large models.
We pre-train SPLAT on English Wikipedia. Specifically we use the KILT Wikipedia snapshot$\footnote{\url{https://huggingface.co/datasets/kilt_wikipedia}}$ from 2019 \citep{petroni2021kilt} as provided by the HuggingFace Datasets library \citep{lhoest2021datasets}.

For both SGD and MultiWOZ we set the shared target values $T$ as the $\texttt{[NONE]}$ and $\texttt{[DONTCARE]}$ tokens and include a special intent with the name "NONE" for each service which is used as the target intent when no other intent is active. We set the maximum answer length $L_\text{ans}$ to 30 tokens.

All experiments are conducted on a machine with eight A100 80GB GPUs. A single training run takes around 12 hours for the base model and 1.5 days for the large model.

\begin{table*}[t]
\centering
\begin{adjustbox}{max width=\textwidth}
\begin{tabular}{@{}lcccc@{}}
\toprule
Model & Pretrained Model & Single-Pass & Intent & JGA \\ \midrule
\emph{With system action annotations} & & & \\ %\midrule
MT-BERT \citep{kapelonis2022multi}           & BERT-base (110M)         & \xmark & 94.7 & 82.7 \\ % not directly comparable
paDST \citep{ma2019end}                      & XLNet-large (340M)       & \xmark & 94.8 & \textbf{86.5} \\ \midrule % not directly comparable
\emph{No additional data} & & & \\ %\midrule
SGD baseline \citep{rastogi2020towards}      & BERT-base (110M)         & \xmark & 90.6 & 25.4 \\
MT-BERT \citep{kapelonis2022multi}   & BERT-base (110M)         & \xmark & -    & 71.9 \\ 
DaP (ind) \citep{lee2021dialogue}            & T5-base (220M)           & \xmark & 90.2 & 71.8 \\
SGP-DST \citep{ruan2020fine}                 & T5-base (220M)           & \xmark & 91.8 & 72.2 \\
D3ST (Base) \citep{zhao2022description}      & T5-base (220M)           & \cmark & 97.2 & 72.9 \\
D3ST (Large) \citep{zhao2022description}     & T5-large (770M)          & \cmark & 97.1 & 80.0 \\
D3ST (XXL) \citep{zhao2022description}       & T5-XXL (11B)             & \cmark & \textbf{98.8} & 86.4 \\ \midrule
SPLAT (Base)                                 & Longformer-base (110M)   & \cmark & 96.7 & 80.1 \\
%Ours (Base) w/ system actions                & Longformer-base (110M)   & \cmark & 96.4 & 83.6 \\ % not pretrained
SPLAT (Large)                                & Longformer-large (340M)  & \cmark & 97.6 & 85.3 \\ \bottomrule
\end{tabular}
\end{adjustbox}
\caption{\label{table:sgd-results}Results on the SGD test set.}
\end{table*}

\begin{table*}[t]
\centering
\begin{adjustbox}{max width=\textwidth}
\begin{tabular}{@{}lcccc@{}}
\toprule
Model & Pretrained Model & Single-Pass & Intent & JGA \\ \midrule
%SGD baseline$\dagger$ \citep{rastogi2020towards}    & \xmark & BERT-base (110M)         & 42.0 \\
%TRADE$\dagger$ \citep{wu2019transferable}            & \xmark & BERT-base (110M)         & 45.4 \\
DS-DST$^\dagger$ \citep{zhang2020find}              & BERT-base (110M)          & \xmark                    & -     & 51.7 \\
Seq2Seq-DU \citep{feng2021sequence}                 & BERT-base (110M)          & \cmark                    & 90.9  & 54.4 \\
LUNA \citep{wang2022luna}                           & BERT-base (110M)          & \xmark                    & -     & 56.1 \\
AG-DST \citep{tian2021amendable}                    & GPT-2 (117M)              & \xmark\rlap{$^\ddagger$}  & -     & 56.1 \\
AG-DST \citep{tian2021amendable}                    & PLATO-2 (310M)            & \xmark\rlap{$^\ddagger$}  & -     & 57.3 \\
DaP (seq) \citep{lee2021dialogue}                   & T5-base (220M)            & \cmark                    & -     & 51.2 \\
DaP (ind) \citep{lee2021dialogue}                   & T5-base (220M)            & \xmark                    & -     & 57.5 \\
D3ST (Base) \citep{zhao2022description}             & T5-base (220M)            & \cmark                    & -     & 56.1 \\
D3ST (Large) \citep{zhao2022description}            & T5-large (770M)           & \cmark                    & -     & 54.2 \\
D3ST (XXL) \citep{zhao2022description}              & T5-XXL (11B)              & \cmark                    & -     & \textbf{58.7} \\ \midrule
SPLAT (Base)                                        & Longformer-base (110M)    & \cmark                    & 91.4  & 56.6 \\
SPLAT (Large)                                       & Longformer-large (340M)   & \cmark                    & \textbf{91.5}  & 57.4 \\ \bottomrule
\end{tabular}
\end{adjustbox}
\caption{\label{table:multiwoz-results}Results on the MultiWOZ 2.2 test set. Results denoted by $\dagger$ were reported in the original MultiWOZ 2.2 paper \citep{multiwoz2.2}. $\ddagger$: AG-DST uses a fixed two-pass generation procedure.}
\end{table*}

\section{Evaluation}

\begin{table*}[t]
\centering
\begin{adjustbox}{max width=\textwidth}
\begin{tabular}{@{}lccccc@{}}
\toprule
Model & Params. & Orig. & Avg. ($v_1$--$v_5$) & Avg. $\Delta$ & Max $\Delta$ \\ \midrule
DaP (ind) \citep{lee2021dialogue}            & 220M & 71.8 & 64.0 & -7.8 & - \\
SGP-DST \citep{ruan2020fine}                 & 220M & 72.2 / 60.5$^*$ & 49.9$^*$ & -10.6 & - \\
D3ST (Large) \citep{zhao2022description}     & 770M & 80.0 & 75.3 & -4.7 & -10.9 \\
D3ST (XXL) \citep{zhao2022description}       & 11B  & \textbf{86.4} & 77.8 & -8.6 & -17.5 \\ \midrule
SPLAT (Base)                            & 110M & 80.1 & 76.0 & -4.1 & -7.8 \\
SPLAT (Large)                           & 340M & 85.3 & \textbf{82.8} & \textbf{-2.5} & \textbf{-5.3} \\ \bottomrule
\end{tabular}
\end{adjustbox}
\caption{\label{table:sgdx-results}Joint goal accuracy on the five different SGD-X schema variants. Results denoted by $^*$ are based on a reimplementation in the SGD-X paper which could not reproduce the original results.}
\end{table*}

\begin{table}[t]
\centering
\begin{adjustbox}{max width=\linewidth}
\begin{tabular}{@{}lcccc@{}}
\toprule
Model & Params. & Seen & Unseen & Overall \\ \midrule
SGP-DST$^1$                 & 220M & 88.0  & 67.0  & 72.2  \\
D3ST (Base)$^2$             & 220M & 92.5  & 66.4  & 72.9  \\
D3ST (Large)$^2$            & 770M & 93.8  & 75.4  & 80.0  \\
D3ST (XXL)$^2$              & 11B  & \textbf{95.8}  & \textbf{83.3}  & \textbf{86.4}  \\ \midrule
SPLAT (Base)                & 110M & 94.5  & 75.2  & 80.1  \\
SPLAT (Large)               & 340M & 94.6  & 82.2  & 85.3  \\ \bottomrule
\end{tabular}
\end{adjustbox}
\caption{\label{table:generalization}Joint goal accuracy on the SGD test set on seen and unseen services. Baseline results are reported by $^1$\citet{ruan2020fine} and $^2$\citet{zhao2022description} respectively.}
\end{table}

We evaluate the effectiveness of our model through a series of experiments designed to answer the following questions: 1) How effective is the proposed model architecture at DST in general? 2) Does the model generalize well to unseen services? 3) Is the model robust to changes in schema such as different slot names and descriptions? 4) Which parts of the model contribute most to its performance?

\subsection{Baselines}

We compare our model to various discriminative and generative baseline approaches. Note that not all of them are directly comparable due to differences in their experimental setups. 

\paragraph{Extractive baselines.}
SGD baseline \citep{rastogi2020towards} is a simple extractive BERT-based model which encodes the schema and last utterance separately and uses the embeddings in downstream classifiers to predict relative slot updates for the current turn. SGP-DST \citep{ruan2020fine} and DS-DST \citep{zhang2020find} are similar but jointly encode utterance and slot schema. Multi-Task BERT \citep{kapelonis2022multi} is also similar but uses system action annotations which include annotations of slots offered or requested by the system (e.g. ``\texttt{[ACTION] Offer [SLOT] location [VALUE] Fremont}''). paDST \citep{ma2019end} combines an extractive component for non-categorical slots with a classifier that uses 83 hand-crafted features (including system action annotations) for categorical slots. Additionally it augments training data via back-translation achieving strong results but making a direct comparison difficult. LUNA \citep{wang2022luna} separately encodes dialogue history, slots and slot values and learns to first predict the correct utterance to condition the slot value prediction on. 

\paragraph{Generative baselines.}
Seq2Seq-DU \citep{feng2021sequence} first separately encodes utterance and schema and then conditions the decoder on the cross-attended utterance and schema embeddings. The decoder generates a state representation consisting of pointers to schema elements and utterance tokens. AG-DST \citep{tian2021amendable} takes as input the previous state and the current turn and learns to generate the new state in a first pass and correcting mistakes in a second generation pass. AG-DST does not condition generation on the schema and slot semantics are learned implicitly so it is unclear how well AG-DST transfers to new services. DaP \citep{lee2021dialogue} comes in two variants which we denote as DaP (seq) and DaP (ind). DaP (ind) takes as input the entire dialogue history and an individual slot description and decodes the inferred slot value directly but requires one inference pass for each slot in the schema. DaP (seq) instead takes as input the dialogue history and the sequence of all slot descriptions and decodes all inferred slot values in a single pass. D3ST \citep{zhao2022description} takes a similar approach and decodes the entire dialogue state including the active intent in a single pass. Categorical slot values are predicted via an index-picking mechanism.

\subsection{Main Results}

\paragraph{Schema-Guided Dialogue.} Table \ref{table:sgd-results} shows results on the SGD test set. We report results for intent accuracy and JGA. We find that our model significantly outperforms models of comparable size in terms of JGA. In particular our 110M parameter SPLAT base model outperforms the 220M model D3ST base model by 7.2 JGA points and even achieves comparable performance to the much larger D3ST large model. Going from SPLAT base to SPLAT large we observe a significant performance improvement. In particular SPLAT large outperforms the D3ST large model by 5.3 JGA and nearly achieves comparable performance to the more than 30$\times$ larger D3ST XXL model.

We note that although paDST achieves the best performance of all baseline models in terms of JGA, it is not directly comparable because it is trained with hand-crafted features and additional back-translation data for training which has been shown to significantly improve robustness and generalization to unseen descriptions in schema-guided DST \citep{sgdx}. Similarly, although Multi-Task BERT achieves good performance this can mostly be attributed to the use of system action annotation as \citet{kapelonis2022multi} themselves demonstrate. Without system action annotations its performance drops to 71.9 JGA.

In terms of intent accuracy SPLAT base slightly underperforms D3ST base and D3ST large by 0.5 and 0.4 JGA while SPLAT large achieves better performance and slightly improves upon the D3ST large performance. Overall, SPLAT achieves strong performance on SGD.

\paragraph{MultiWOZ.} Table \ref{table:multiwoz-results} shows results on the MultiWOZ 2.2 test set. As the majority of papers does not report intent accuracy on MultiWOZ 2.2 we focus our analysis on JGA. We find that SPLAT base outperforms most similarly-sized models including D3ST base and large and that SPLAT large performs better than all models aside from the more than 30$\times$ larger D3ST XXL. The notable exceptions to this are AG-DST and DaP (ind). AG-DST large achieves performance that is similar to SPLAT large using a generative approach but it performs two decoding passes, employs a negative sampling strategy to focus on more difficult examples and is trained for a fixed schema. DaP (ind) also achieves similar performance but needs one inference pass for every slot at every turn of the dialogue. This is much slower and simply not realistic in real-world scenarios with a large number of available services and slots. The sequential variant DaP (seq) which instead outputs the full state in a single pass performs much worse.

\paragraph{Comparison.} While DaP (ind) shows strong performance that matches SPLAT on MultiWOZ, SPLAT fares much better than DaP (ind) on the SGD dataset. This can be seen to be indicative of a stronger generalization ability as MultiWOZ uses the same schema at training and test time whereas SGD includes new, unseen services at test time and thus requires the model to generalize and understand the natural language schema descriptions.

\subsection{Robustness}
DST models which take natural language descriptions of intents and slots as input naturally may be sensitive to changes in these descriptions. In order to evaluate the robustness of our model to such linguistic variations we perform experiments on the SGD-X benchmark. The SGD-X benchmark comes with five crowd-sourced schema variants $v_1$ to $v_5$ which increasingly diverge in style from the original schema. We train SPLAT on SGD and evaluate it on the test set using all five different schema variants.

As shown in Table~\ref{table:sgdx-results}, our model is considerably more robust to linguistic variations than all of the baseline models. On average SPLAT base loses around 4.1 points and SPLAT large loses around 2.5 points joint goal accuracy when compared to the results on the original schema. When considering the mean performance across all unseen schema variants SPLAT large significantly outperforms the more than 30$\times$ larger D3ST XXL by 5.0 points. These observations also hold for the base model: the 110M parameter SPLAT base even outperforms the 11B parameter D3ST XXL on the least similar schema variant $v_5$ further highlighting the superior robustness of our model.

\subsection{Generalization to unseen domains}
In real-world scenarios virtual assistants cover a wide range of services that can change over time as new services get added or removed requiring dialogue models to be re-trained. One of our goals is to improve generalization to unseen services thus minimizing the need for expensive data collection and frequent re-training. As the MultiWOZ dataset does not include any new and unseen services in its test set our analysis primarily focuses on the SGD dataset. Table~\ref{table:generalization} shows results on SGD with a separate evaluation for dialogues in seen and unseen domains. We find that SPLAT achieves better generalization and improves upon the baselines with a particularly large margin on unseen domains where SPLAT base outperforms D3ST base by 8.8 points and SPLAT base outperforms D3ST large by 6.8 points.

\begin{table}[t]
\centering
\begin{adjustbox}{max width=\linewidth}
\begin{tabular}{@{}lccccc@{}}
\toprule
& & \multicolumn{2}{c}{SGD} & \multicolumn{2}{c}{MultiWOZ} \\ \cmidrule(l){3-4} \cmidrule(l){5-6}
Model & Params. & Intent & JGA & Intent & JGA \\ \midrule
Longformer (extr.) & 110M & 95.9 & 78.5 & \textbf{91.4} & 55.5 \\
+ SPM & 110M & \textbf{97.0} & 79.0 & \textbf{91.4} & 56.1 \\
+ SPM + RSS-PT & 110M & 96.7 & \textbf{80.1} & \textbf{91.4} & \textbf{56.6} \\ \midrule
Longformer (extr.) & 340M & 97.5 & 83.5 & 91.4 & 56.3 \\
+ SPM & 340M & \textbf{98.2} & 83.8 & 91.4 & \textbf{57.8} \\
+ SPM + RSS-PT & 340M & 97.6 & \textbf{85.3} & \textbf{91.5} & 57.4 \\ \bottomrule
\end{tabular}
\end{adjustbox}
\caption{\label{table:ablation-results}Ablation results on the SGD and MultiWOZ test sets. Longformer (extr.) refers to an extractive model with no span representations and simple start and end pointers for answer prediction, SPM refers to the Span Pointer Module and RSS-PT to pre-training with the Recurrent Span Selection objective.}
\end{table}

\subsection{Ablation Study}
We conduct an ablation study to identify the contribution of the different components to model performance. Results can be seen in Table \ref{table:ablation-results}. We compare a variant of our model which does not use span representations (referred to as ``Longformer (extractive)'') but instead has two pointers \texttt{[SLOT]} and \texttt{[/SLOT]} which are used to select the start and end of the answer span. We find that using the Span Pointer Module to directly select the span improves performance across both model sizes and datasets. Furthermore, we find pre-training our model for better span representations via the recurrent span selection task to be crucial giving further significant performance gains for all sizes and datasets except the 340M parameter model on the MultiWOZ dataset where JGA slightly deteriorates. Across both model sizes gains from RSS pre-training are larger on the SGD dataset. We hypothesize that this may be attributed to better span representations learned through RSS pre-training which in turn generalize better to unseen domains.

\section{Related Work}

\paragraph{Extractive DST.} Following the traditional extractive setting \citet{chao2019bertdst} propose a machine reading comprehension (MRC) approach which decodes slot values turn-by-turn using a different learned classifier for each slot. As a classifier has to be learned for each new slot this approach cannot easily be transferred to new slots.

Schema-guided approaches address this by explicitly conditioning predictions on a variable schema which describes intents and slots in natural language \citep{rastogi2020towards}. Both \citet{ruan2020fine} and \citet{zhang2021sgd} introduce schema-guided models but predict slots independently from one another requiring multiple encoder passes for each turn and failing to model intent-slot and inter-slot dependencies. \citet{ma2019end} use MRC for non-categorical and handcrafted features for categorical slots.

\paragraph{Generative DST.}

In an attempt to address the lack of ability to generalize to new domains and ontologies, \citet{wu2019transferable} propose incorporating a generative component into DST. Based on the dialog history and a domain-slot pair a state generator decodes a value for each slot. However as each slot is decoded independently the approach cannot model slot interdependencies. \citet{feng2021sequence} instead generate the entire state as a single sequence of pointers to the dialogue history and input schema but separately encode history and schema. \citet{zhao2021effective} model DST fully as a text-to-text problem and directly generate the entire current state as a string. \citet{lin2021zero} transfer a language model fine-tuned for seq2seq question answering to DST zero-shot using the dialog history as context and simply asking the model for the slot values. By also including a natural language schema in the input, \citet{zhao2022description} show that full joint modeling and rich attention between history and schema lead to better results in DST. Furthermore, they demonstrate the flexibility of this fully language driven paradigm by leveraging strong pre-trained language models for cross-domain zero-shot transfer to unseen domains. \citet{gupta2022show} show the effectiveness of using demonstrations of slots being used in practice instead of a natural language descriptions in the prompt.

\section{Conclusion}
In this work we introduced SPLAT, a novel architecture for schema-guided dialogue state tracking which learns to infer slots by learning to select target spans based on natural language descriptions of slot semantics, and further showed how to pre-train SPLAT via a recurrent span selection objective for better span representations and a stronger slot prediction performance. We find that our proposed architecture yields significant improvements over existing models and achieving 85.3 JGA on the SGD dataset and 57.4 JGA on the MultiWOZ dataset. In schema-guided DST the ability to generalize to new schemas and robustness to changes in schema descriptions is of particular interest. We demonstrated that our model is much more robust to such changes in experiments on the SGD-X benchmark where SPLAT outperforms the more than 30$\times$ larger D3ST-XXL model by 5.0 points.

\section*{Limitations}

One trade-off of limiting the prediction space using an extractive pointer module is that it does not support prediction of multiple slot values which is necessary for some dialogues in the MultiWOZ 2.3 and 2.4 datasets. To keep the architecture simple we do not consider cases in which slots take multiple values in this work, but we can effectively adapt our model for this setting by introducing sequential query tokens for each slot. Another limitation is that the span representation requires a computation of $O(N \cdot L_\text{ans})$ complexity where $N$ and $L_\text{ans}$ represent the length of context and answer span, respectively. For very long answers this might occur significant computational costs compared to existing span prediction approaches which have $O(N)$ complexity. However, this can be alleviated by adding a simple sampling and filtering step during training and prediction. We plan to further study and address these limitations in future work. 

\section*{Ethics Statement}
We introduced a novel model architecture for schema-guided dialogue state tracking which leverages a natural language schema and a span pointer module to achieve higher accuracy in dialogue state tracking. All experiments were conducted on publicly available datasets which are commonly used in research on dialogue systems.

%\section*{Acknowledgements}

\bibliography{references}
\bibliographystyle{acl_natbib}

\newpage
\FloatBarrier

\appendix

\section{Appendix}
\label{sec:appendix}

\begin{table}[ht]
    \begin{tabular}{@{}ll@{}}
        \toprule
        Symbol   & Definition    \\ \midrule
        $\texttt{LAT}$ & Linear Attention Transformer \\
        $\mathcal{I}$  & Input sequence \\
        $\mathcal{G} $ & Global inputs \\
        $\mathcal{M}$ & Set of masked recurring span clusters \\
        $\mathcal{R}$ & Set of all recurring span clusters \\
        $D^\text{intent}$ & Intent descriptions \\
        $D^\text{slot}$ & Intent descriptions \\
        $E$ & Joint encoding obtained from \texttt{LAT} \\
        $I$ & Intents \\
        $S$ & Slots \\
        $T$ & Shared target tokens \\
        $U$ & Utterances \\
        $\mathbf{h}^\texttt{[INTENT]}$ & Intent embedding \\
        $\mathbf{h}^\texttt{[SLOT]}$ & Slot embedding \\
        $\mathbf{h}^\texttt{[UTT]}$ & Utterance embedding \\
        $\mathbf{h}_{ij}^\texttt{SPAN}$ & Span embedding from position $i$ to $j$ \\
        $\mathbf{x}_i$ & Token representation at position $i$ \\
        $\theta$ & Model parameters \\
        \bottomrule
    \end{tabular}
    \caption{Glossary of symbols}
\end{table}

\end{document}